\documentclass[preprint,12pt]{elsarticle}

\usepackage{lineno}
\usepackage{amssymb}
\usepackage[ruled,lined]{algorithm2e}
\usepackage{amsmath}
\usepackage{amsfonts}
\usepackage{amssymb}
\usepackage{graphicx}
\usepackage{caption}
\usepackage{float}
\usepackage[font=footnotesize,labelfont=bf]{subcaption}
\newcommand{\argmin}{\mathop{\mathrm{arg\,min}}}

\begin{document}

\begin{frontmatter}

\title{Enhancing Fruit and Vegetable Detection in Unconstrained Environment with a Novel Dataset}

\author[inst1]{Sandeep Khanna}

\affiliation[inst1]{organization={Department of Computer Science and Engineering},
            addressline={Indian Institute of Technology Jodhpur}}

\author[inst2]{Chiranjoy Chattopadhyay}
\author[inst1]{Suman Kundu}

\affiliation[inst2]{organization={School of Computing and Data Sciences},
            addressline={FLAME University}}

\begin{abstract}
 Automating the detection of fruits and vegetables using computer vision is essential for modernizing agriculture, improving efficiency, ensuring food quality, and contributing to technologically advanced and sustainable farming practices. This paper presents an end-to-end pipeline for detecting and localizing fruits and vegetables in real-world scenarios. To achieve this, we have curated a dataset named FRUVEG67 that includes images of 67 classes of fruits and vegetables captured in unconstrained scenarios, with only a few manually annotated samples per class. We have developed a semi-supervised data annotation algorithm (SSDA) that generates bounding boxes for objects to label the remaining non-annotated images. For detection, we introduce the Fruit and Vegetable Detection Network (FVDNet), an ensemble version of YOLOv7 featuring three distinct grid configurations. We employ an averaging approach for bounding-box prediction and a voting mechanism for class prediction. We have integrated Jensen-Shannon divergence (JSD) in conjunction with focal loss to better detect smaller objects. Our experimental results highlight the superiority of FVDNet compared to previous versions of YOLO, showcasing remarkable improvements in detection and localization performance. We achieved an impressive mean average precision (mAP) score of 0.78 across all classes. Furthermore, we evaluated the efficacy of FVDNet using open-category refrigerator images, where it demonstrates promising results.
\end{abstract}


\end{frontmatter}


\section{Introduction}
\label{intro}
Fruit and vegetable detection is essential for various agriculture applications, such as yield estimation, quality classification, harvest automation, and food safety. However, detecting fruits and vegetables in unconstrained environments, such as outdoor orchards or markets, poses significant challenges due to varying illumination, occlusion, clutter, and diversity of shapes, sizes, and colors. There has been tremendous work done in the computer vision community for object detection and localization \cite{zhao2019object} for constrained settings. However, relatively less work is done on object detection in unconstrained environments \cite{xiao2022few,chen2018domain,peng2020large}. 
YOLOv7, a recent object detection model \cite{wang2022yolov7}, has achieved state-of-the-art results for the MS-COCO dataset \cite{lin2014microsoft}, which comprises 80 classes. It is important to note that this dataset is prepared primarily from images captured in a constrained scenario. 

Although there are many publicly available data sets for object detection, the availability of datasets for modernizing agriculture is limited. In addition, labeling such images poses several unique challenges compared to labeling images in controlled settings. Unlike controlled environments where there may be established visual references or markers for labeling, such images often lack such references. In addition, such images can be complex and contain multiple objects, occlusions, and background clutter. Annotators must carefully delineate and label each object accurately, taking into account their boundaries, poses, and variations.

\begin{figure}
    \centering
    \subfloat[\centering Sample 1]{{\includegraphics[width=3.8cm, height=2.2cm]{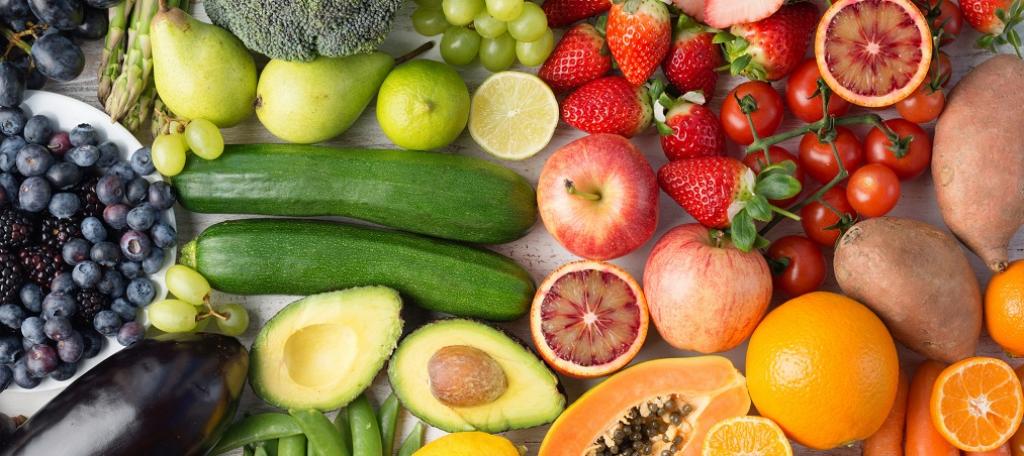} }}%
    \qquad
    \subfloat[\centering Sample 2]{{\includegraphics[width=3.8cm, height=2.2cm]{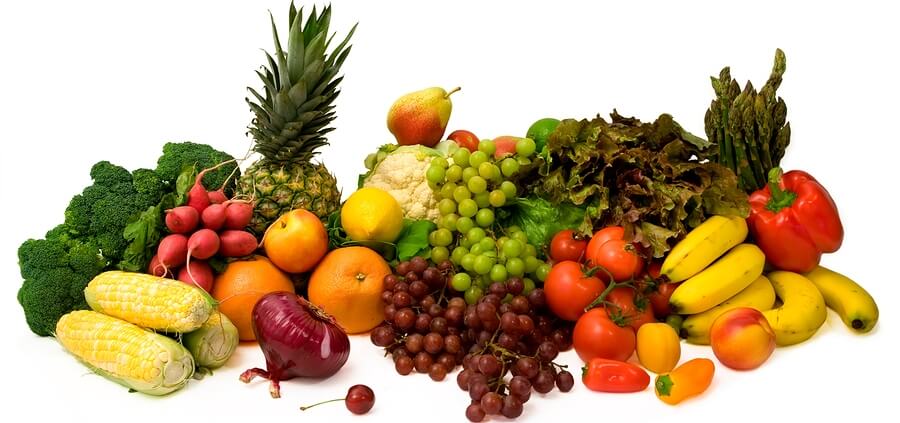} }}%
    \qquad
    \subfloat[\centering Sample 3]{{\includegraphics[width=3.8cm, height=2.2cm]{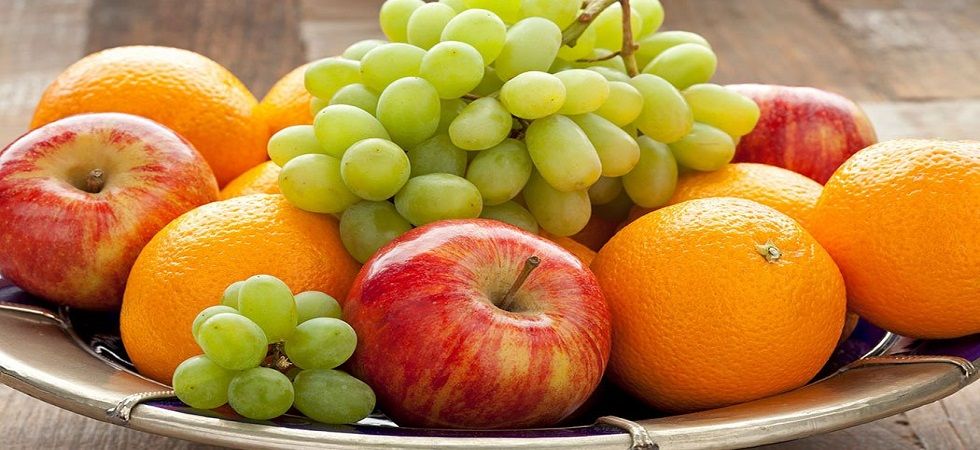} }}%
    \qquad
    \subfloat[\centering Sample 4]{{\includegraphics[width=3.8cm, height=2.2cm]{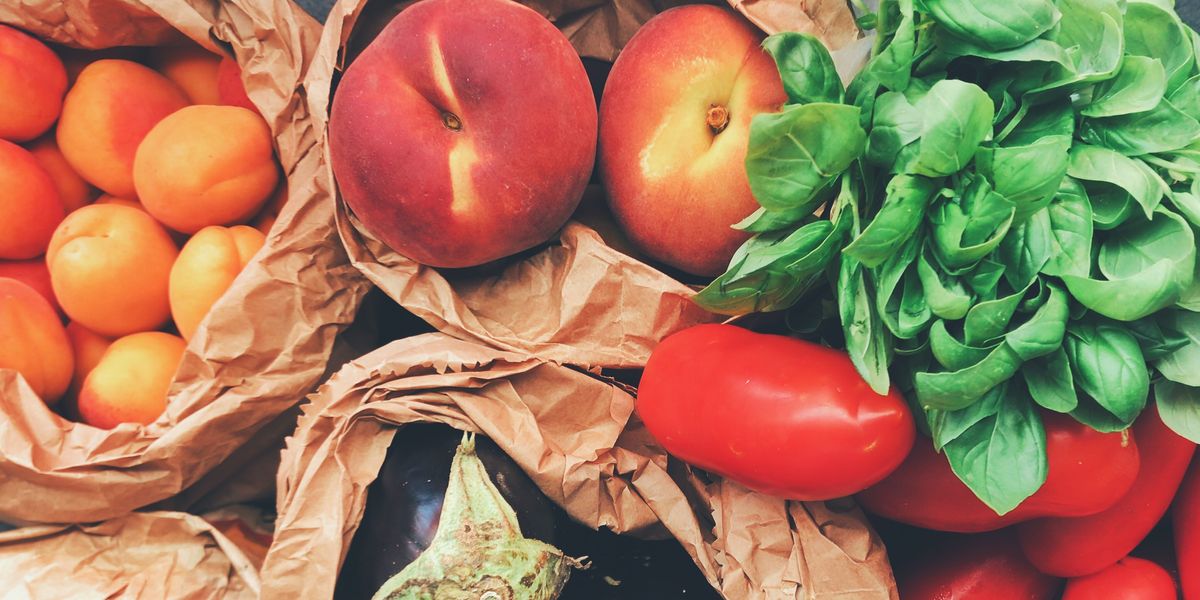} }}%
    \caption{Sample Images of the proposed dataset.}%
    \label{fig:example2}%
\end{figure}

\begin{figure}[!htbp]
\centering
\includegraphics[width=0.8\linewidth]{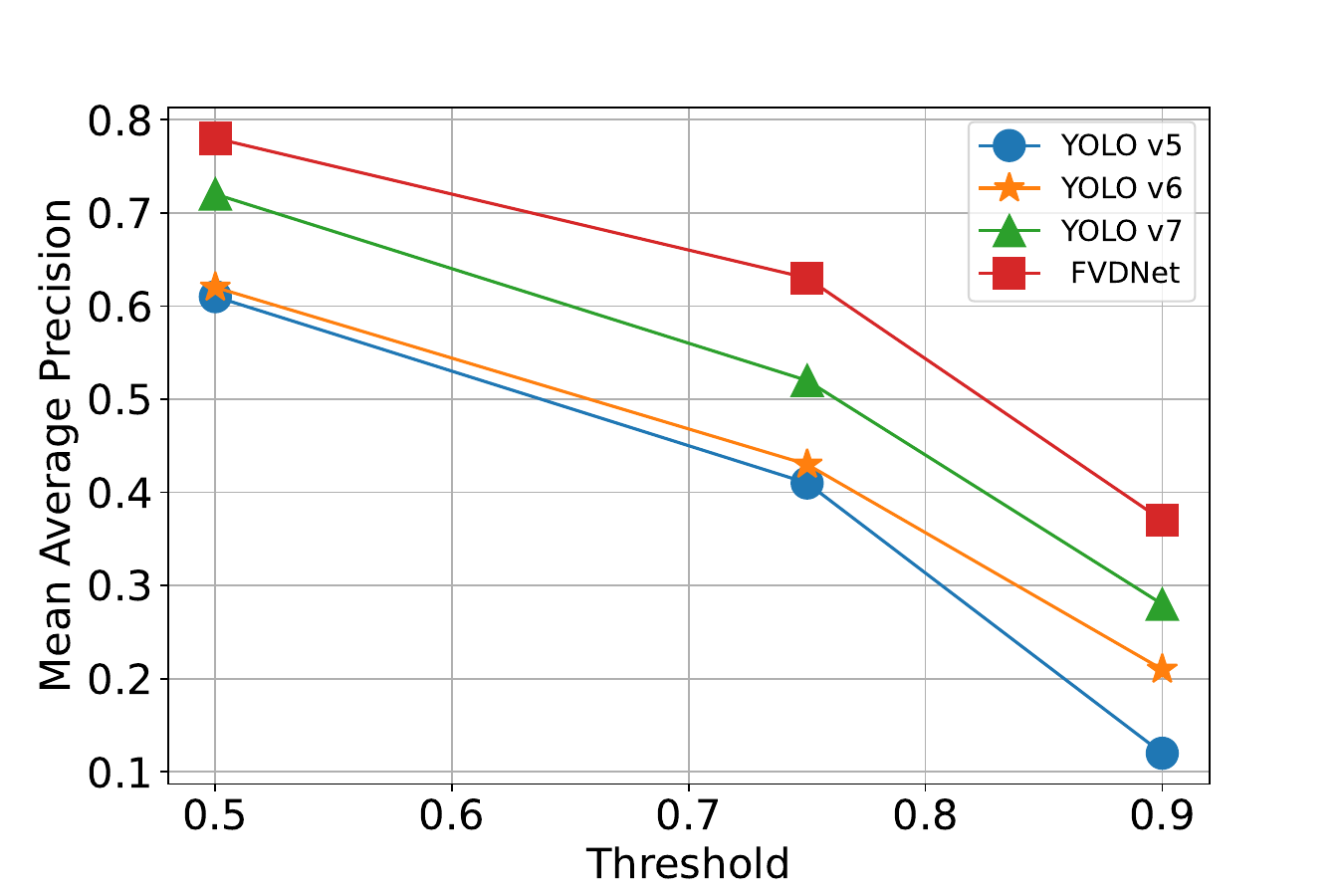}
\caption{Mean Average Precision vs. Threshold Comparison.}
\label{fig:example28}
\end{figure}

In this study, we created and prepared the FRUVEG67 dataset, which encompasses 67 different classes of fruits and vegetables. Figure \ref{fig:example2} displays a selection of sample images from the dataset. We collected 5000 images, dividing them between 35 categories of vegetables and 32 categories of fruits. Around 2000 images were manually annotated. For the remaining, we've proposed a semi-supervised learning algorithm (SSDA) for generating object annotations. SSDA runs iteratively to annotate objects in images with few samples learned on YOLOv7. The major challenge of detecting objects in unconstrained scenarios is occlusion; sometimes, the object size is too small for the model to capture finer details. Also, often, objects are cluttered with excessive noise. Our approach is based on three paradigms: pre-processing images, proposing a model, and redefining loss function.
First, the pre-processing pipeline for images in FRUVEG67 has been designed. Additionally, we have introduced Fruit and Vegetable Detection Network (FVDNet), an ensemble variant of YOLOv7 that incorporates three unique grid configurations with sizes of 32, 16, and 8. We employ a novel approach for bounding box prediction and a voting mechanism for class prediction. We have integrated Jensen-Shannon Divergence (JSD) \cite{menendez1997jensen} in conjunction with focal loss, enabling accurate object detection even for tiny objects. We made a comparison between FVDNet and prior YOLO versions. The mean average precision (mAP) results for different thresholds (0.5, 0.75, 0.9) are presented in Figure \ref{fig:example28}.

\begin{figure}[!b]
    \centering
    \subfloat[\centering Sample 1]{{\includegraphics[width=3.8cm, height=2.2cm]{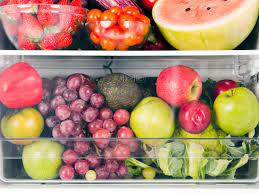} }}%
    \qquad
    \subfloat[\centering Sample 2]{{\includegraphics[width=3.8cm, height=2.2cm]{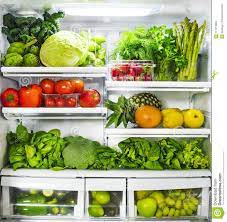} }}%
    \qquad
    \subfloat[\centering Sample 3]{{\includegraphics[width=3.8cm, height=2.2cm]{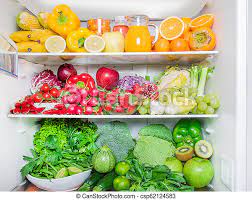} }}%
    \qquad
    \subfloat[\centering Sample 4]{{\includegraphics[width=3.8cm, height=2.2cm]{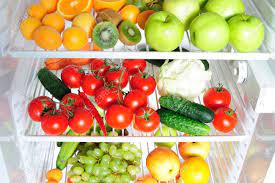} }}%
    \caption{Sample open category Images from Refrigerator.}%
    \label{fig:example19}%
\end{figure}

As shown in the Figure, FVDNet consistently outperforms the other YOLO models at all thresholds. Furthermore, ablation investigations on FRUVEG67 and VOC Dataset 2012 \cite{pascal-voc-2012} were shown by altering different backbone networks paired with one-stage, two-stage, and transformer-based detectors.
Also, we have tested FVDNet with Kullback-Leibler Divergence (KLD) embedded with the focal loss. Subsequently, we assessed the proposed model's performance by evaluating open-category images obtained from a refrigerator. Sample images from this evaluation are illustrated in Figure \ref{fig:example19}. 

The following were the major contributions. 
\begin{enumerate}
    \item Creation and proposal of FRUVEG67 dataset.
    \item Design of SSDA for annotating objects images.
    \item Proposed FVDNet model for detection and localization of fruits and vegetables.
    \item Incorporated JSD loss for object localization as the difference in Gaussian distributions.
    \item Case study presented on images captured from Refrigerator.
\end{enumerate}

The rest of the paper is structured as follows:
In Section \ref{sec:rel}, we presented the literature review. Section \ref{sec:dat} introduces the dataset with collection and preparation. Section \ref{sec:meth} discusses the overall methodology, i.e., the pipeline for carrying out the tasks, covering automatic annotations of objects, preprocessing, FVDNet and JSD loss. Section \ref{sec:expres} shows the experiments, results and ablation studies for all the above-defined tasks. Finally, in Section \ref{sec:conclu}, we present the conclusion and future work. 

\section{LITERATURE REVIEW}
\label{sec:rel}
In this section, we review the existing methods and techniques for detecting objects (specifically fruits and vegetables) in unconstrained environments and highlight their advantages and limitations. Below, we detail cutting-edge research in Object identification and localization and Ensemble Learning.

\paragraph{Object Detection and Localization}
Detection and localization of objects are broadly classified into three types: one-stage detectors, two-stage detectors, and transformer-based detectors \citep{liu2022image,bi2022srrv,liu2022petr}.  YOLO and Fully Convolutional One-stage Object Detection (FCOS) are the primary foundations for the most cutting-edge real-time object detectors. Early research on object recognition was based on template matching techniques and simple part-based models \cite{fischler1973representation}. However, deeper CNNs have led to record-breaking improvements in detecting more general object categories. This shift came about when the successful application of DCNNs in image classification \cite{krizhevsky2017imagenet}  was transferred to object detection, resulting in the milestone Region-based CNN (RCNN)  detector of \cite{girshick2014rich}. 

Redmon et al. \cite{redmon2016you} proposed YOLO, a unified detector casting object detection as a regression problem from image pixels to spatially separated bounding boxes and associated class probabilities. YOLOv2 and YOLO9000 \cite{redmon2017yolo9000} proposed YOLOv2, an improved version of YOLO, in which the custom GoogLeNet \cite{szegedy2015going} network is replaced with the simpler DarkNet19, plus batch normalization. In a later stage, the authors proposed YOLOv3 \cite{redmon2018yolov3}. It has two points: using multi-scale features for object detection and adjusting the basic network structure. YOLOv4 \cite{bochkovskiy2020yolov4} style has a significant change, more focus on comparing data, and substantially improved. Li et al. \cite{LI2022107418} propose an approach for powdery mildew on strawberry leaves. However, the latest release of the YOLOv7 \cite{wang2022yolov7} model has created a benchmark and surpasses all known object detectors in speed and accuracy. Liu et al. \cite{LIU2023107834} proposed a real-time dynamic system for fruit detection and localization. \cite{zhu2024detection} have tried to identify maturity of multi-cultivar olive fruit using object detection model. 
On a similar note \cite{gharaghani2020ripeness} have tried to map ripeness of orange fruit using object detection approach. The authors \cite{mim2018automatic} have developed an end-end pipeline for automatic detection of mango ripening stages based on object detection approach.
Similarly \cite{khojastehnazhand2019maturity} have applied image processing technique to for volume estimation of apricot. We have employed the YOLOv7 model to generate annotations and created FVDNet for object detection and localization of fruits and vegetables in unconstrained environment.

\paragraph{Ensemble Learning:}
Ensemble learning is widely acknowledged for achieving highly accurate predictions \mbox{\cite{dong2020survey}}. It can be broadly categorized into two main approaches: bagging and boosting. A genetic algorithm-based ensemble of deep CNN methods was proposed by \cite{AYAN2020105809} for crop pest classification. In \cite{9558994}, ensemble learning methods were proposed for Alzheimer's Disease detection, showing that the AdaBoost ensemble method improved the classification rate from 3.2\% to 7.2\%. In \cite{xu2021forest}, a forest fire detection using ensemble learning was proposed. In \cite{ASTANI2022107054}, a tomato disease classification approach was proposed. They integrated Yolov5 and EfficientDet models and observed a performance increase of 2.5\% to 10.9\% in fire detection accuracy. An ensemble pre-processing approach was proposed for paddy-moisture online detection in \cite{YAN2022107050}. In \cite{TALUKDER2023100155}, authors have proposed a robust Deep Ensemble Convolutional Neural Network (DECNN) model that can accurately diagnose rice nutrient deficiency.

Overall, these studies highlight the efficacy of ensemble learning methods in various domains. In our approach, we have utilized the bagging ensemble method to enhance the accuracy of final predictions for images. Based on the literature review, we have identified a research gap in the availability of a large and diverse dataset of fruit and vegetable images captured in different scenarios and locations and a need for a novel deep learning-based framework that can enhance detection and classification performance in complex settings. We present our proposed dataset and framework to address this gap in the next section.

\section{{Dataset of Fruits and Vegetables (FRUVEG67)}}
\label{sec:dat}
FRUVEG67 is a dataset comprises of $67$ categories of fruits ($34$) and vegetables ($33$). A detailed description of dataset collection and preparation is defined in the sub-sections below.

\subsection{Dataset Collection}
\begin{figure}[t]
\centering
\includegraphics[width=\linewidth]{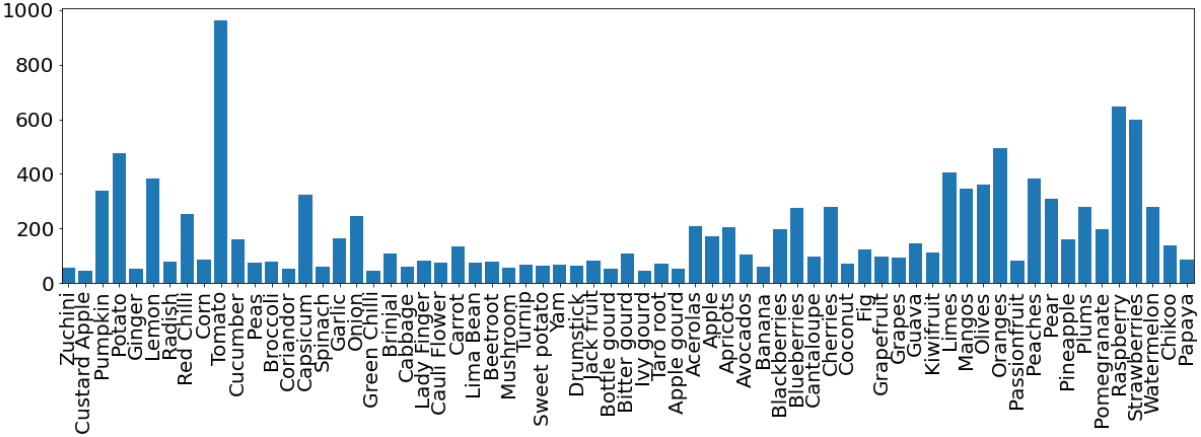}
\caption{Data Distribution across 67 classes of FRUVEG67.}
\label{fig:example55}
\end{figure}

The images of fruits and vegetables in various unconstrained scenarios were collected using the Flickr API. Apart from the images, some individual images are gathered for each category so that the model can learn features specific to a particular category more refinedly. A total of 67 different classes are collected. Figure \ref{fig:example2} shows sample images of FRUVEG67. Images that do not contain either fruits or vegetables were removed using ResNet-52 \cite{he2016deep} trained on the fruits and vegetables data set. After filtering, we are left with $5000$ images of unconstrained ($3500$) and individual ($1500$) categories. Figure \ref{fig:example55} shows the overall data distribution across all classes.

\subsection{Data Preparation}
For data preparation, using LabelImg \cite{tzutalin2015labelimg}, we have annotated $2000$ images manually, considering that each category must be annotated at least $20$. Some of the images contain more than $14$ categories. As far as we know, this is the first-ever data set generated comprising fruits and vegetable images in such a scenario with $67$ categories.

\section{Methodology}
\label{sec:meth}
\begin{figure*}[!b]
\centering
\includegraphics[width=\linewidth]{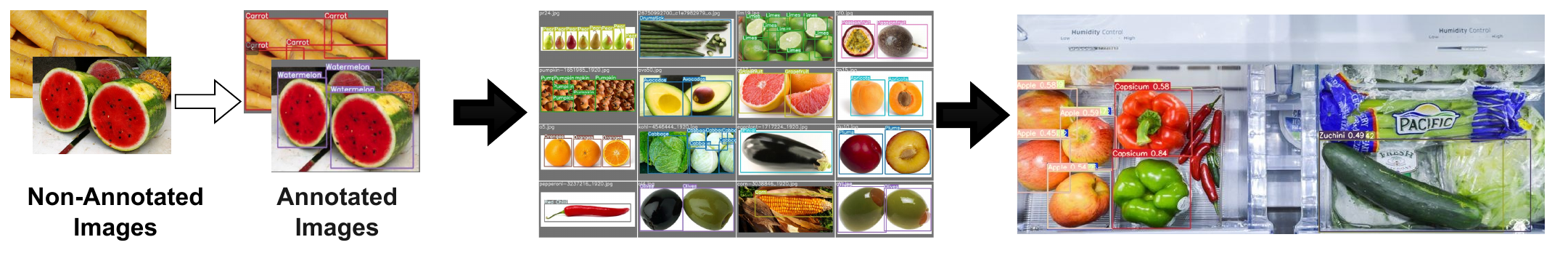}
\caption{Block Diagram of the proposed methodology: It comprises of two significant steps. Step 1: a semi-supervised method for annotating unlabeled images, Step 2: objects in images are detected. As a downstream task detection on open category images captured from a refrigerator is demonstrated.}
\label{fig:3}
\end{figure*}

This paper's overall approach (sequence of tasks) is depicted in Figure \ref{fig:3}. In the following subsections, we present them in detail.

\subsection{Semi-Supervised Data Annotation (SSDA) Algorithm}
\label{sec:alg}
The proposed semi-supervised data annotation algorithm, defined in Algorithm \ref{alg:1}, takes the annotated images in $Train\_Set$ and the non-annotated images in $Test\_Set$ as inputs. $model$ used is YOLOv7 in our case and $\theta = 4$ is the maximum number of iterations for the algorithm to finish all annotations. The output of SSDA will be  $final\_Train\_Set$ and $final\_Test\_Set$. For the first iteration the $Train\_Set$ were fed into YOLOv7 model pre-trained on Microsoft’s Common Objects in Context (MS-COCO) dataset comprising of 80 classes. Once the model is trained then $Test\_Set$ were inferred on the trained model. The images with maximum objects having confidence score $\geq 0.5$ are added to the train set. At the same time those images were removed from the test set as they have been annotated by model. The model is fine-tuned in each subsequent iteration. The process is repeated until iteration count $< \theta$. Finally when iteration count exceeds $\theta$ then the remaining non-annotated images are the ones that definitely require human annotations. These images were manually annotated then added to the original train set of first iteration. This is how we obtained the final train and test sets (Line no. \ref{ssda:op}). More details on SSDA is provided in supplementary file.

\begin{algorithm}[!htb]
\footnotesize
\KwIn{$Train\_Set , Test\_Set, model, \theta$}
\KwOut{$final\_Train\_Set , final\_Test\_Set$}

\nl $add\_set = \{\}$\;
\nl $newTrain\_Set = Train\_Set$\;
\nl $newTest\_Set = Test\_Set$\;
\nl \For{$i = 0$ \KwTo $\theta$}{       \label{ssda:1}
\nl Train model on $newTrain\_Set$\;
\nl Test model on $newTest\_Set$\;
\nl \For{$i = 0$ \KwTo len($Test\_Set$)}{
\nl \If{conf\_score of maximum object in $Test\_Set[i] \geq 0.5$}{
\nl $add\_set = add\_set \cup Test\_Set[i]$\; 
}}
\nl $newTrain\_Set = newTrain\_Set \cup add\_set$\;
\nl $newTest\_Set = newTest\_Set \setminus add\_set$\;
\nl $SSDA(newTrain\_Set, newTest\_Set, model, \theta - 1)$\;                                   \label{ssda:2}
}
\nl \For{$i = 0$ \KwTo len($newTest\_Set$)}{
\nl \If{conf\_score of maximum object in $newTest\_Set[i] \geq 0.5$}{
\nl $add\_set = add\_set \cup newTest\_Set[i]$\;
}
\nl $finalTrain\_Set = Train\_Set \cup add\_set$\;
\nl $finalTest\_Set = Test\_Set \setminus add\_set$\;
}
\nl \Return $finalTrain\_Set$, $finalTest\_Set$ \label{ssda:op}
\caption{$SSDA (Train\_Set, Test\_Set, model, \theta)$} \label{alg:ssd}
\label{alg:1}
\end{algorithm}

\subsection{Pre-processing}
Given our primary focus on handling images captured in unconstrained scenarios featuring obscured objects with varying sizes and lighting and reduced transparency, we implemented a series of pre-processing steps on the dataset images. The aim was to standardize image sizes and reduce computational complexity by resizing images to the YOLOv7 default dimension of $640 \times 640$. Normalization was applied to adjust pixel values to a standardized range, enhancing model performance and convergence. The mean and standard deviation values used for normalization were [0.485, 0.456, 0.406] and [0.229, 0.224, 0.225], respectively, typical for models trained on the ImageNet dataset.
To reduce noise, Gaussian Smoothing was applied, and histogram equalization was employed to enhance object visibility. Additionally, the dataset underwent a scaling transformation to provide the model with multi-scale image features. Multi-scale features involve resizing the image while maintaining the same aspect ratio. This process facilitates the model in learning to detect objects of varying scales, making smaller objects more prominent and easier to detect.


\subsection{FVDNet}
\label{sec:yoloarch}
\begin{figure}[t]
\centering
\includegraphics[width=\linewidth]{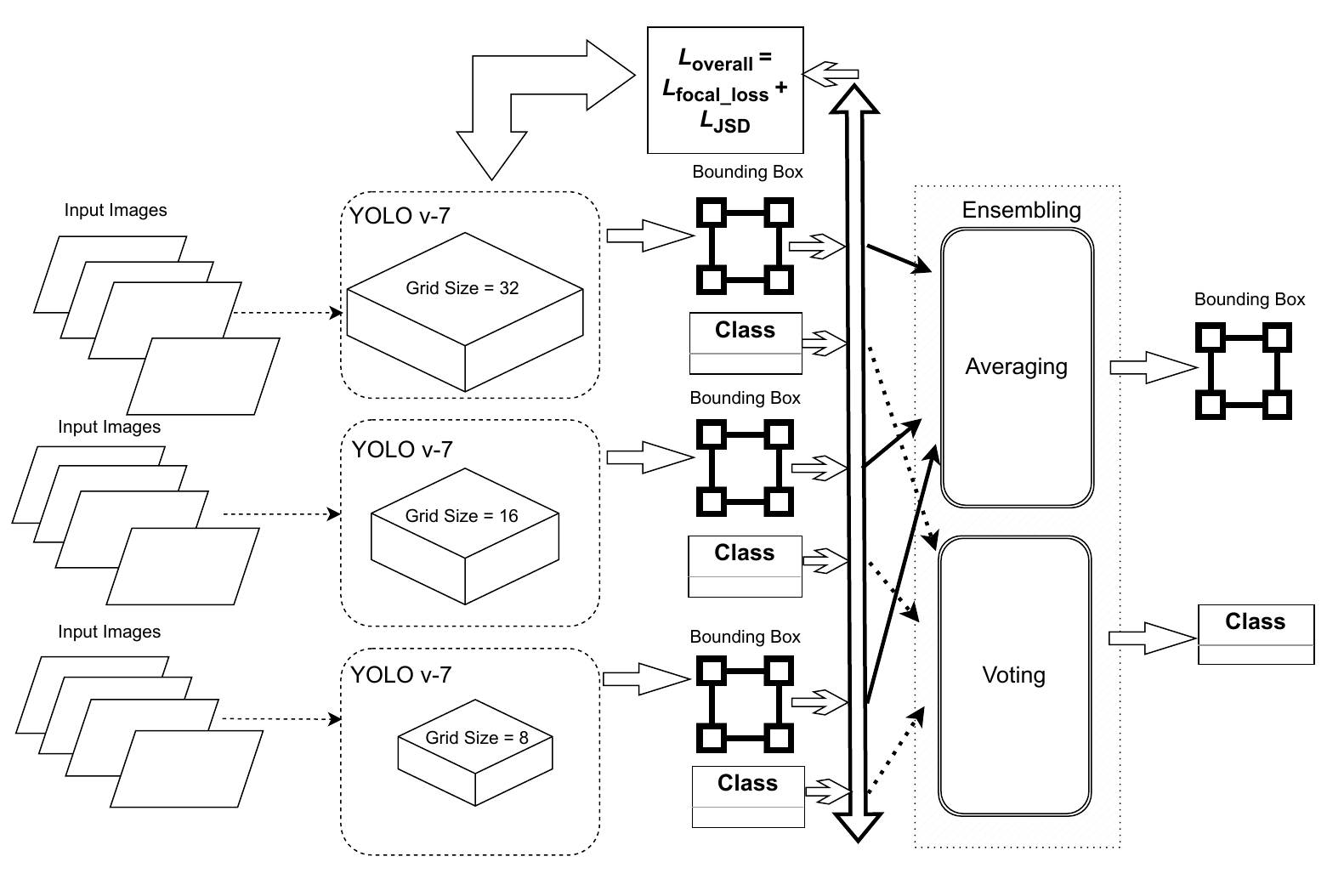}
\caption{An illustration on the proposed FVDNet model that consists of three parallel training processes, where YOLOv7 models are trained with distinct configurations. 
}
\label{fig:15}
\end{figure}

We harness the capabilities of ensemble learning to enhance accuracy, employing a combination of three YOLOv7 models, each with unique configurations. To address the challenge of effectively detecting smaller objects, we tailor the grid size for each YOLOv7 model. While one model uses the default grid size of $32$, we progressively reduce it to $16$ and $8$ in the other configurations. Throughout the training process, we meticulously ensure that all models maintain visibility and accuracy in detecting objects of varying sizes. We were able to capture a wide range of object sizes in the FRUVEG67 dataset using an ensemble of YOLOv7 models with varied grid sizes. By leveraging ensemble learning and adapting the grid size for each YOLOv7 model, FVDNet can accurately identify objects of various dimensions and positions, even amidst cluttered environment. This leads to a significant improvement in overall detection performance, increasing the precision and recall of our system. Figure \ref{fig:15} depicts our proposed FVDNet model.

After completing the training process, the final predictions were derived by employing a fusion of three YOLOv7 models. For bounding box regression, we leverage an averaging approach, wherein the regression outputs provided by the three models are combined. This aggregation technique serves a crucial purpose in refining and enhancing the accuracy of bounding box predictions. Furthermore, to determine the final class prediction, we adopt a voting system. By pooling together the class predictions from all three models, we identify the class that receives the highest number of votes, which becomes the final class prediction. The utilization of an ensemble of multiple models and aggregation techniques has been widely recognized as an effective strategy to enhance predictive accuracy and reduce the risk of overfitting \cite{7780459, zhang2021vit, azevedo2020stochastic}. By combining regression outputs and utilizing a voting system for class predictions, our ensemble approach leverages the strengths and diversity of individual models, enhancing the overall prediction's robustness and reliability. This method reduces potential errors or biases in a single model, leading to improved performance and generalization capabilities in the prediction system.

We opted for a combination of Jensen-Shannon Divergence (JSD) and focal loss, which have proven effective in related research. JSD, commonly utilized as a similarity measure between probability distributions, complements the focal loss function, an extension of the cross-entropy loss, designed to prioritize hard negatives during training. Empirically, we observed that this ensemble-based approach, incorporating JSD, significantly boosts the average precision (AP) value across a majority of classes. This result aligns with findings in previous studies where JSD has been applied to improve model performance in various tasks \cite{Lin2017}. This ensemble technique, combined with the inclusion of JSD, allows us to leverage the diverse capabilities of the models and address the challenges posed by object detection tasks effectively.

\subsection{Modeling bounding box offset as a Gaussian distribution}
Instead of minimizing the loss function of bounding box in the form of regression, the loss function is adjusted to minimize the distribution loss based on the calculation of $\mu$ (mean), and $\sigma$ (standard deviation) of single variate Gaussian on the target distribution of $x, y, w,$ and $h$ coordinate. The $\mu$ represents the center of the bounding box, and the $\sigma$ represents the uncertainty or variability in the prediction. Likewise the target bounding box are mapped with $\mu$ and $\sigma$ values. 

We have calculated $\sigma$ as a fraction of the bounding box dimensions (width and height). This approach allows the model to have higher tolerance for larger objects and lower tolerance for smaller objects. $\sigma$ is defined as

\begin{equation}
    \sigma = k \times \max(width, height)
\end{equation}

Here $k$ is a constant scaling learnable factor. To represent the bounding box coordinates as Gaussian distributions, we have calculated the PDF (probability density function) values for each coordinate using the predicted mean and standard deviation \cite{He_2019_CVPR}. Below equations shows the calculation of predicted $P(k)$ and ground truth $Q(k)$ for each $k \in \{x, y, w, h\}$.

\begin{equation}
   P(k) = \frac{1}{{\sigma_p \sqrt {2\pi } }}e^{{{ - \left( {k - \mu_p } \right)^2 } \mathord{\left/ {\vphantom {{ - \left( {k - \mu_p } \right)^2 } {2\sigma_p ^2 }}} \right. \kern-\nulldelimiterspace} {2\sigma_p ^2 }}}
\end{equation}

\begin{equation}
   Q(k) = \frac{1}{{\sigma_{gt} \sqrt {2\pi } }}e^{{{ - \left( {k - \mu_{gt} } \right)^2 } \mathord{\left/ {\vphantom {{ - \left( {k - \mu_{gt} } \right)^2 } {2\sigma_{gt} ^2 }}} \right. \kern-\nulldelimiterspace} {2\sigma_{gt} ^2 }}}
\end{equation}

here $\mu_p$, $\mu_{gt}$, $\sigma_p$ and $\sigma_{gt}$ are the mean of predicted, mean of ground-truth, standard deviation of predicted and ground-truth respectively.
When $\sigma_p$ = 0 , it means the model is extremely confident about estimated bounding box location.

\subsection{Jenson Shanon Divergence as a Similarity Measure}
The Jensen-Shannon Divergence (JSD) is typically used as a similarity measure between probability distributions. JSD is used as a similarity metric between the predicted object distribution and the ground truth distribution. By comparing the distributions, one can assess how well the predicted bounding box aligns with the ground truth bounding box. This concept is useful for evaluating the quality of object localization, especially for small objects. One important property of the Jensen-Shannon divergence \cite{nielsen2020generalization} is that it is symmetric, meaning that $JSD(P || Q) = JSD(Q || P)$ and this Jensen-Shannon distance is always bounded. Here $P$ is the target probability distribution and $Q$ is the distribution predicted by the model. This symmetry property of JSD considers weighted average of KL divergence from both the distribution. Therefore accounting for a more balanced measure of divergence between distributions, and can be effective in guiding the model towards aligning the predicted and ground truth distributions in object detection tasks. The expression of JSD is defined as:

\begin{equation}
    \mathcal{L}_{JSD}(P,Q) = 0.5 \times \mathcal{KL} (P || M) + 0.5 \times \mathcal{KL} (Q || M)
\end{equation}

Here, $M = 0.5 \times (P + Q)$ is the average distribution computed, the JSD loss combines the KL divergence of both P and Q from the average distribution M. The KL divergence from the predicted distribution P to the ground truth distribution Q is computed using:

\begin{equation}
    \mathcal{KL}(P || Q) = \int P(x) log\left(\frac{P(x)}{Q(x)}\right)dx
\end{equation}

Here $\mathcal{KL}(P || Q)$ represents the Kullback-Leibler Divergence between distributions $P$ and $Q$ and x is any uni-variate random variable.

We have computed the JSD loss for each coordinate (x, y, width, height) separately and sum up the losses for all coordinates to obtain the total JSD loss. The goal of the object localization is to estimate the $\mathcal{\theta}$ that minimizes the below objective function.

\begin{equation}
    {\theta}^* = \argmin_\theta \mathcal{L}_{JSD}(P,Q)
\end{equation}

\subsection{Overall Loss Function}
   
The overall loss function is modified to

\begin{equation}
    \mathcal{L}_{overall} = \mathcal{L}_{focal\_loss} + \mathcal{L}_{JSD}
\end{equation}

where $\mathcal{L}_{overall}$ is the overall loss, $\mathcal{L}_{focal\_loss}$ is the focal loss and $\mathcal{L}_{JSD}$ is the Jensen-Shannon Divergence loss.

\begin{equation}
    \mathcal{L}_{focal\_loss}(p_t) = - \alpha_t (1 - p_t)^\gamma \log \log(p_t) 
\end{equation}

where $(1 - p_t)^\gamma$ is the cross entropy loss, with a tunable focusing parameter $\gamma \geq 0$. We have experimented with five values of gamma ranging from (0, 0.5, 1, 2, 5). $\alpha$ is the balanced variant of the focal loss and $p_t \in [0,1]$, is the model’s estimated probability for the class. 

\section{Experiments \& Results}
\label{sec:expres}
We conducted experiments using FVDNet on the FRUVEG67 dataset, using various configurations. We conducted a comparative analysis between the outcomes of FVDNet and earlier cutting-edge iterations of YOLO. Furthermore, we assessed the effectiveness of our suggested model by using open category photos obtained from a refrigerator. We conduct ablation investigations using the FRUVEG67 and Pascal VOC 2012 datasets, which consist of 20 categories of objects. The purpose of these research is to assess the efficacy of our model by employing alternative backbone networks on a range of state-of-the-art models. Furthermore, we used KLD instead of JSD as the loss function to assess its influence on the final outcomes. The detailed analysis of these data is discussed in the following sub-sections.

\subsection{FVDNet}
\begin{figure*}[t]
    \centering
    \subfloat[\centering YOLOv5 Prediction]{{\includegraphics[width=6.2cm, height=3.2cm]{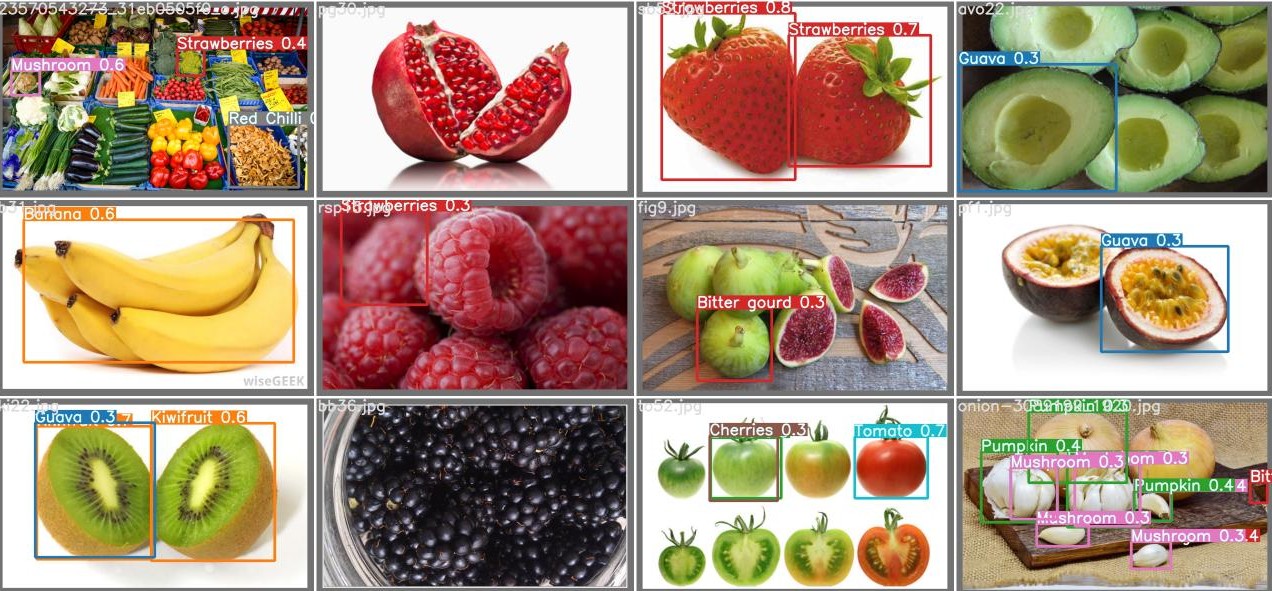} }}%
    \qquad
    \subfloat[\centering YOLOv6 Prediction]{{\includegraphics[width=6.2cm, height=3.2cm]{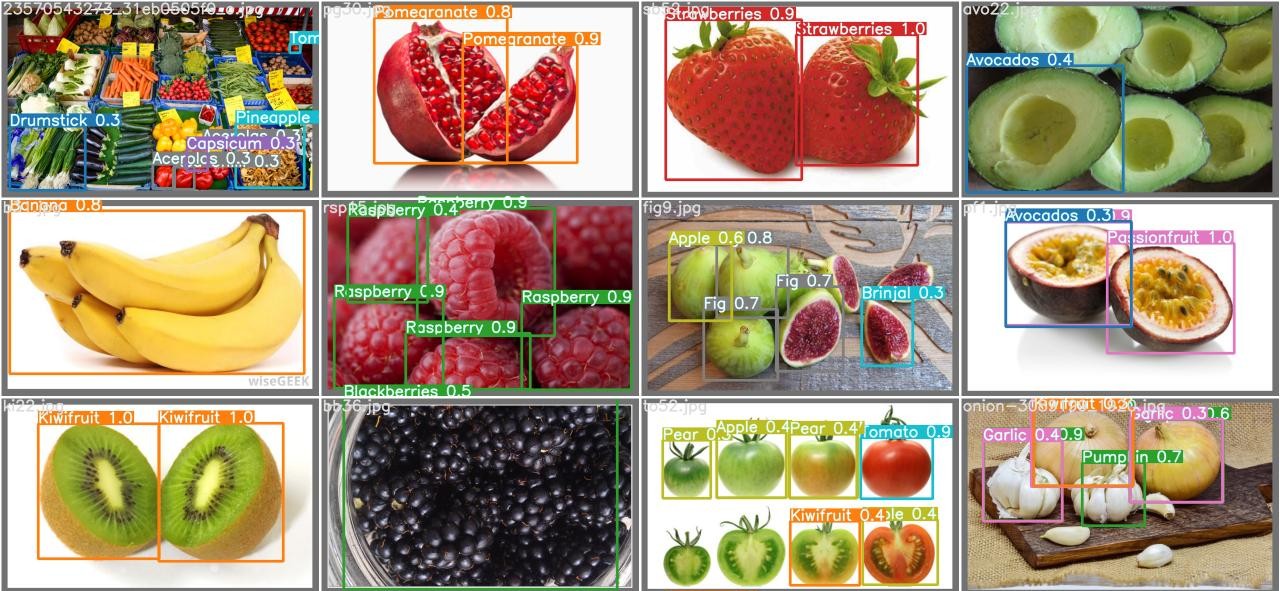} }}%
    \qquad
    \subfloat[\centering YOLOv7 Prediction]{{\includegraphics[width=6.2cm, height=3.2cm]{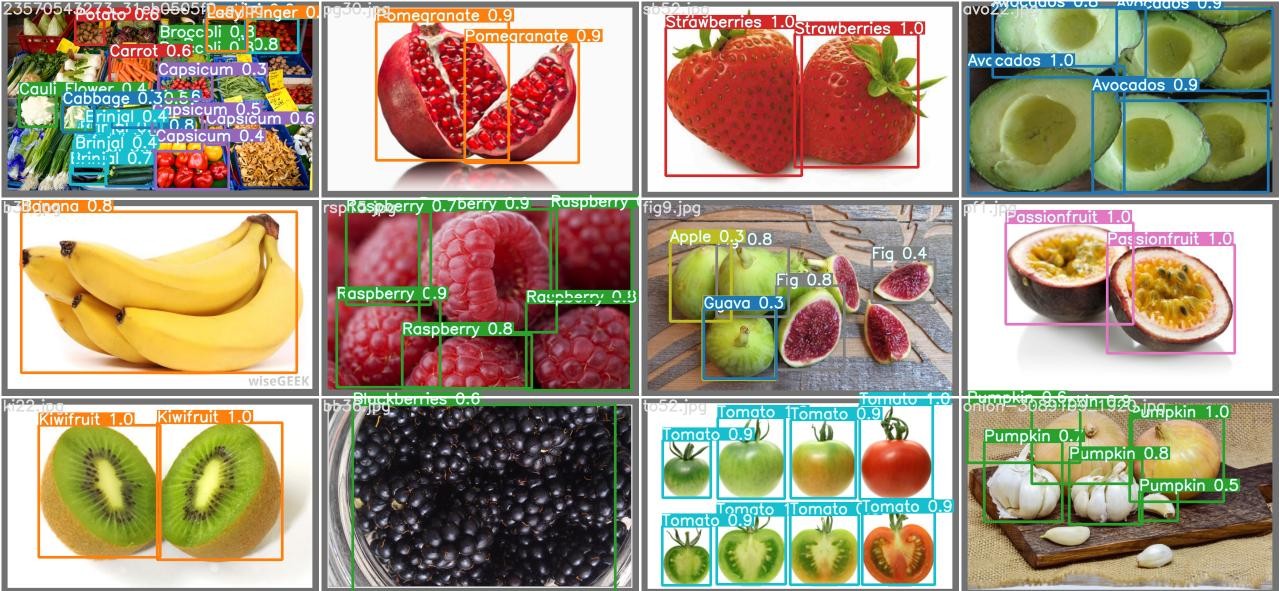} }}%
    \qquad
    \subfloat[\centering FVDNet Prediction]{{\includegraphics[width=6.2cm, height=3.2cm]{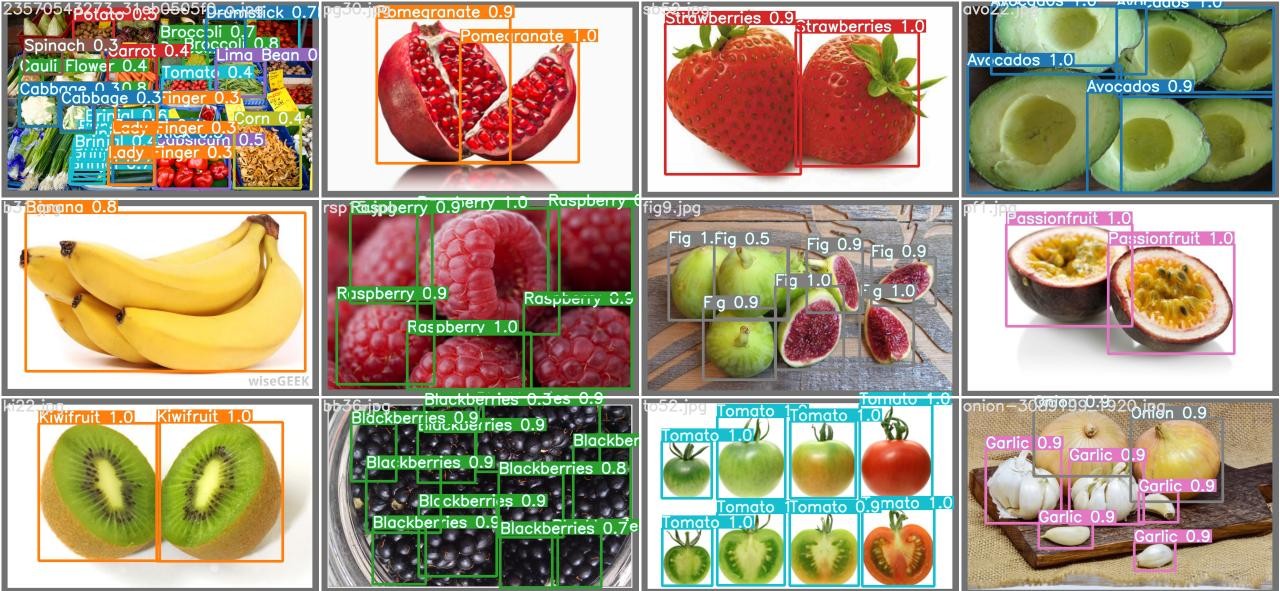} }}%
    \caption{Visualization of Results on YOLO v5, v6, v7 and FVDNet: Overall, FVDNet outperforms in both localization and detection of objects.
    }%
    \label{fig:example7}%
\end{figure*}

The model was trained for 100 epochs using two NVIDIA A30 GPUs. The dataset was divided into 70\% for training, 20\% for testing, and 10\% for validation. The results presented in Figure \ref{fig:example7} demonstrate the performance of various models, including YOLO v5, v6, v7, and FVDNet, when tested on different images. Notably, FVDNet proves to be highly effective in detecting objects of all sizes, even in scenarios where objects are clustered and overlapping. It stands out by surpassing human performance in identifying parts of objects that were overlooked by both humans and SSDA. Furthermore, the model demonstrates its capability to recognize and accurately localize even small sections of objects.

We conducted experiments using three different threshold values (0.5, 0.75, and 0.9) for three YOLO predecessor models as well as our own model. The results are depicted in Figure \ref{fig:example28}. The graph clearly illustrates that the FVDNet consistently outperforms the existing models across all threshold values in terms of mean average precision (mAP). 

Additionally, we performed a comparison of the average precision (AP) for all 67 classes, specifically for a threshold of 0.5. The results are presented in Table \ref{tab:Tab1}. The AP values were computed based on the summation of confidence scores provided by the model for each category, divided by the total number of instances for that particular object across all test images. The precision was calculated as:

\begin{equation}
    AP_{class} = \frac{\sum_{i \in I} \sum_{class \in i} {ConfidenceScore_{class}}}{\sum_{i \in I} \sum_{class \in i} {Sum_{class}}} 
\end{equation}

here $i$ represents an index for the image set $I$, $class$ represents class of an object and $Sum_{class}$ represents the number of samples belonging to a given class.

FVDNet demonstrated superior performance compared to its previous versions. The tomato class exhibit the highest mAP of $0.94$ followed by watermelon and Strawberries. Most classes achieved an average precision (AP) greater than or equal to $0.5$. However, certain classes had lower AP scores, primarily due to the limited number of instances of those objects present in the images.

The mean average precision of the entire model is reported to be $0.78$ and is calculated using the below equation.

\begin{equation}
    mAP_{model} = \frac{\sum_{class \in C}{AP_{class}}} {C}
\end{equation}

here $C$ represents the total number of classes. In our case $C=67$.

\begin{table*}[!htbp]
\centering
\caption{Class-wise Average Precision(AP) of FVDNet with existing YOLO (v5 (Y5), v6 (Y6) and v7 (Y7)) models on FRUVEG67 dataset}
\resizebox{\textwidth}{!}{
\begin{tabular}{lllllllllllllll}
\hline
\textbf{Models/Class}                                                   & \textbf{Y5} & \textbf{Y6} & \textbf{Y7} & \textbf{\begin{tabular}[c]{@{}l@{}}FVDNet \end{tabular}} & \textbf{Models/Class}                                                 & \textbf{Y5} & \textbf{Y6} & \textbf{Y7} & \textbf{\begin{tabular}[c]{@{}l@{}}FVDNet \end{tabular}} & \textbf{Models/Class}       & \textbf{Y5} & \textbf{Y6} & \textbf{Y7} & \textbf{\begin{tabular}[c]{@{}l@{}}FVDNet \end{tabular}} \\ \hline

\textbf{Zuchini}                                                  & 0.20        & 0.47        & 0.53        & 0.65                                                        & \textbf{Carrot}                                                 & 0.41        & 0.69        & 0.74        & 0.75                                                        & \textbf{Fig}          & 0.16        & 0.35        & 0.44        & 0.61                                                        \\ \hline
\textbf{\begin{tabular}[c]{@{}l@{}}Custard \\ Apple\end{tabular}} & 0.15        & 0.60        & 0.61        & 0.64                                                        & \textbf{Lima Bean}                                              & 0.12        & 0.36        & 0.51        & 0.54                                                        & \textbf{Grapefruit}   & 0.14        & 0.32        & 0.31        & 0.42                                                        \\ \hline
\textbf{Pumpkin}                                                  & 0.35        & 0.62        & 0.63        & 0.72                                                        & \textbf{Beetroot}                                               & 0.17        & 0.51        & 0.53        & 0.62                                                        & \textbf{Grapes}       & 0.31        & 0.55        & 0.59        & 0.69                                                        \\ \hline
\textbf{Potato}                                                   & 0.52        & 0.67        & 0.74        & 0.89                                                        & \textbf{Mushroom}                                               & 0.09        & 0.21        & 0.19        & 0.44                                                        & \textbf{Guava}        & 0.24        & 0.51        & 0.63        & 0.70                                                        \\ \hline
\textbf{Ginger}                                                   & 0.22        & 0.55        & 0.47        & 0.56                                                        & \textbf{Turnip}                                                 & 0.15        & 0.32        & 0.28        & 0.46                                                        & \textbf{Kiwifruit}    & 0.26        & 0.43        & 0.47        & 0.63                                                        \\ \hline
\textbf{Lemon}                                                    & 0.61        & 0.66        & 0.75        & 0.84                                                        & \textbf{Sweet potato}                                           & 0.13        & 0.24        & 0.26        & 0.49                                                        & \textbf{Limes}        & 0.24        & 0.47        & 0.45        & 0.55                                                        \\ \hline
\textbf{Radish}                                                   & 0.17        & 0.41        & 0.43        & 0.57                                                        & \textbf{Yam}                                                    & 0.17        & 0.33        & 0.32        & 0.54                                                        & \textbf{Mangoes}      & 0.27        & 0.63        & 0.66        & 0.67                                                        \\ \hline
\textbf{Red Chilli}                                               & 0.19        & 0.57        & 0.56        & 0.62                                                        & \textbf{Drumstick}                                              & 0.23        & 0.34        & 0.38        & 0.57                                                        & \textbf{Olives}       & 0.22        & 0.47        & 0.49        & 0.53                                                        \\ \hline
\textbf{Corn}                                                     & 0.21        & 0.36        & 0.32        & 0.52                                                        & \textbf{Jack fruit}                                             & 0.31        & 0.51        & 0.57        & 0.67                                                        & \textbf{Oranges}      & 0.63        & 0.74        & 0.81        & 0.89                                                        \\ \hline
\textbf{Tomato}                                                   & 0.74       & 0.77        & 0.81        & 0.94                                                       & \textbf{Bottle gourd}                                           & 0.18        & 0.41        & 0.39        & 0.62                                                        & \textbf{Passionfruit} & 0.17        & 0.51        & 0.54        & 0.58                                                        \\ \hline
\textbf{Cucumber}                                                 & 0.36        & 0.65        & 0.66        & 0.70                                                        & \textbf{Bitter gourd}                                           & 0.19        & 0.44        & 0.46        & 0.63                                                        & \textbf{Peaches}      & 0.19        & 0.52        & 0.55        & 0.62                                                        \\ \hline
\textbf{Peas}                                                     & 0.26        & 0.48        & 0.51        & 0.56                                                        & \textbf{Taro root}                                              & 0.13        & 0.63        & 0.66        & 0.65                                                        & \textbf{Pear}         & 0.23        & 0.48        & 0.53        & 0.61                                                        \\ \hline
\textbf{Broccoli}                                                 & 0.35        & 0.75        & 0.78        & 0.73                                                        & \textbf{\begin{tabular}[c]{@{}l@{}}Apple \\ gourd\end{tabular}} & 0.21        & 0.49        & 0.62        & 0.57                                                        & \textbf{Pineapple}    & 0.35        & 0.62        & 0.65        & 0.73                                                        \\ \hline
\textbf{Coriandor}                                                & 0.12        & 0.39        & 0.42        & 0.44                                                        & \textbf{Acerolas}                                               & 0.32        & 0.56        & 0.67        & 0.71                                                        & \textbf{Plums}        & 0.21        & 0.55        & 0.62        & 0.52                                                        \\ \hline
\textbf{Capsicum}                                                 & 0.28        & 0.67        & 0.70        & 0.73                                                        & \textbf{Apple}                                                  & 0.52        & 0.71        & 0.72        & 0.78                                                        & \textbf{Pomegranate}  & 0.26        & 0.49        & 0.57        & 0.61                                                        \\ \hline
\textbf{Spinach}                                                  & 0.21        & 0.29        & 0.31        & 0.54                                                        & \textbf{Apricots}                                               & 0.27        & 0.62        & 0.66        & 0.73                                                        & \textbf{Raspberry}    & 0.58        & 0.63        & 0.68        & 0.73                                                        \\ \hline
\textbf{Garlic}                                                   & 0.31        & 0.44        & 0.52        & 0.57                                                        & \textbf{Avocados}                                               & 0.21        & 0.51        & 0.52        & 0.56                                                        & \textbf{Strawberries} & 0.54        & 0.58        & 0.64        & 0.74                                                        \\ \hline
\textbf{Onion}                                                    & 0.35        & 0.67        & 0.66        & 0.71                                                        & \textbf{Banana}                                                 & 0.33        & 0.62        & 0.64        & 0.67                                                        & \textbf{Watermelon}   & 0.23        & 0.66        & 0.68        & 0.78                                                        \\ \hline
\textbf{Green Chilli}                                             & 0.26        & 0.58        & 0.54        & 0.63                                                        & \textbf{Blackberries}                                           & 0.14        & 0.28        & 0.32        & 0.47                                                        & \textbf{Chikoo}       & 0.14        & 0.49        & 0.51        & 0.63                                                        \\ \hline
\textbf{Brinjal}                                                  & 0.37        & 0.52        & 0.53        & 0.68                                                        & \textbf{Blueberries}                                            & 0.17        & 0.27        & 0.31        & 0.46                                                        & \textbf{Papaya}       & 0.31        & 0.56        & 0.61        & 0.68                                                        \\ \hline
\textbf{Cabbage}                                                  & 0.28        & 0.51        & 0.55        & 0.59                                                        & \textbf{Cantaloupe}                                             & 0.13        & 0.37        & 0.43        & 0.52                                                        & \textbf{Ivy gourd}    & 0.11        & 0.48        & 0.42        & 0.47                                                        \\ \hline
\textbf{Lady Finger}                                              & 0.27        & 0.41        & 0.46        & 0.48                                                        & \textbf{Cherries}                                               & 0.20        & 0.51        & 0.49        & 0.59                                                        &                       &             &             &             &                                                             \\ \hline
\textbf{Cauli Flower}                                             & 0.35        & 0.62        & 0.65        & 0.64                                                        & \textbf{Coconut}                                                & 0.21        & 0.49        & 0.57        & 0.62                                                        &                       &             &             &             &                                                             \\ \hline
\end{tabular}
}
\label{tab:Tab1}
\end{table*}

\subsection{Results on Open Category Images from Refrigerator}
\begin{figure}[t]
    \centering
    \subfloat[\centering Sample 1]{{\includegraphics[width=3.8cm, height=2.8cm]{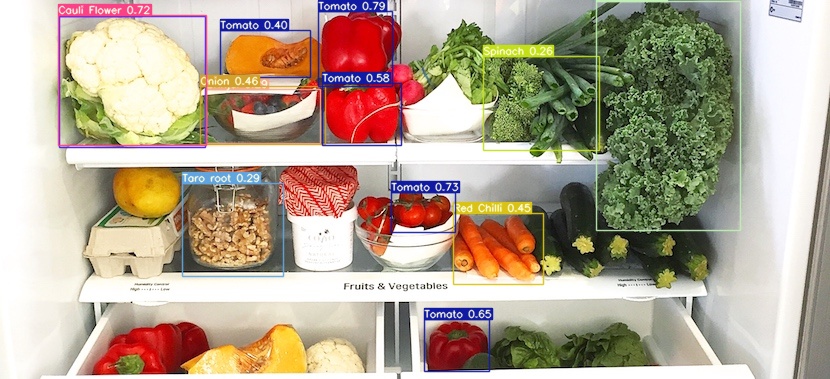} }}%
    \qquad
    \subfloat[\centering Sample 4]{{\includegraphics[width=3.8cm, height=2.8cm]{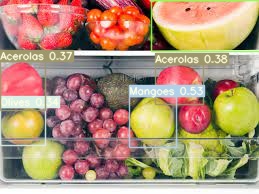} }}%
    \qquad
    \subfloat[\centering Sample 5]{{\includegraphics[width=3.8cm, height=2.8cm]{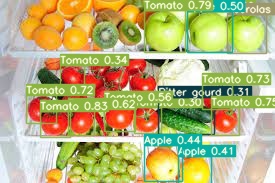} }}%
    \qquad
    \subfloat[\centering Sample 6]{{\includegraphics[width=3.8cm, height=2.8cm]{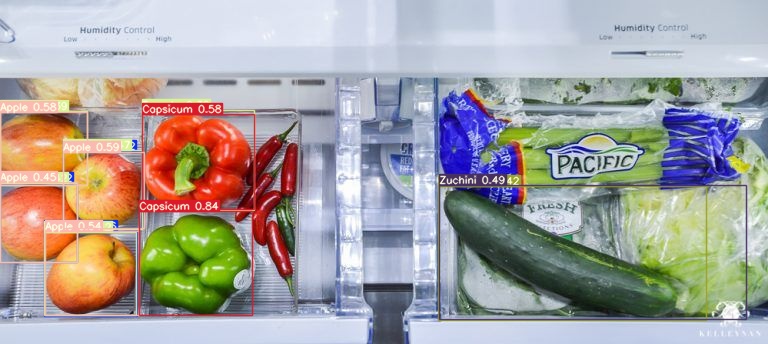} }}%
    \caption{Visualization of FVDNet results on Open Category Images.}%
    \label{fig:example54}%
\end{figure}

We conducted experiments using FVDNet on open images taken from a refrigerator, and the results are illustrated in Figure \ref{fig:example54}. FVDNet shows impressive performance in accurately locating and detecting objects. On carefully examining the last image, we can see that the model successfully distinguished between Zucchini, capsicum, and apples, showcasing its ability to handle multiple objects effectively.

In some cases, the model failed to recognize chillies, likely due to limited samples during training and validation. These findings indicate room for enhancing the model's accuracy and robustness. Nonetheless, FVDNet demonstrated object detection capabilities in refrigerator images, showcasing its potential with scope for improvement.

\section{Discussion}
The subsequent sub-sections are dedicated to analyze the results and findings in more details.

\subsection{Ablation Study}
We have shown the results on FRUVEG67 and VOC 2012. We have experimented by changing the backbone networks, such as RESNET-152 (R-152), DenseNet-169 (D-169), and InceptionNet (IN), to extract image features alongside FVDNet. The outcome of these experiments for FRUVEG67 is presented in the Table \ref{tab:Tab3}, showcasing the results obtained. FVDNet outperforms other models for threshold of 0.5 and 0.9. Specifically, when using a threshold of 0.5, the combination of FVDNet and InceptionNet achieves the highest performance. For a threshold of 0.75, FVDNet with its default backbone network delivers the best results. Finally, when aiming for a threshold of 0.9, utilizing FVDNet in conjunction with R-152 yields the most optimal outcomes. Table \ref{tab:Tab5} shows the results of VOC 2012. Among the evaluated models, Faster R-CNN v2 demonstrated the highest accuracy across all thresholds, while FVDNet closely competed for the second position. This observation suggests that our models perform better on unconstrained data. A possible explanation for this phenomenon could be attributed to the fixed grid size employed in various configurations, which has the potential to result in a reduction in precision.

\begin{table}[!htbp]
\centering
\footnotesize
\caption{Comparison of mAP at different thresholds (0.5, 0.75, 0.9) for Single stage , Multi stage and transformer based detector with FVDNet with different backbone network on FRUVEG67 Dataset.}
\begin{tabular}{lccc}
\hline
\textbf{Models/ mAP}                 & \multicolumn{1}{l}{\textbf{mAP@0.5}} & \multicolumn{1}{l}{\textbf{mAP@0.75}} & \multicolumn{1}{l}{\textbf{mAP@0.90}} \\ \hline
\multicolumn{4}{c}{\textbf{Multi Stage Detector}}                                                                                                           \\ \hline
\textbf{Fast R-CNN}                  & 0.69                                 & 0.49                                  & 0.31                                  \\ \hline
\textbf{Faster R-CNN v2}             & 0.75                                 & \textbf{0.66}                         & 0.35                                  \\ \hline
\multicolumn{4}{c}{\textbf{Single Stage Detector}}                                                                                                          \\ \hline
\textbf{Yolo v5}                     & 0.61                                 & 0.41                                  & 0.12                                  \\ \hline
\textbf{Yolo v6}                     & 0.62                                 & 0.43                                  & 0.21                                  \\ \hline
\textbf{Yolo v7}                     & 0.72                                 & 0.52                                  & 0.28                                  \\ \hline
\textbf{FVDNet}                & \textbf{0.78}                        & 0.63                                  & \textbf{0.37}                         \\ \hline
\textbf{FVDNet + R-152}   & 0.76                                 & 0.56                                  & 0.34                                  \\ \hline
\textbf{FVDNet + D-169} & 0.73                                 & 0.59                                  & 0.36                                  \\ \hline
\textbf{FVDNet + IN} & 0.74                                 & 0.53                                  & 0.32                                  \\ \hline
\multicolumn{4}{c}{\textbf{Transformer based Detector}}                                                                                                     \\ \hline
\textbf{Pix2Seq}                     & 0.69                                 & 0.45                                  & 0.25                                  \\ \hline
\end{tabular}
\label{tab:Tab3}
\end{table}

\begin{table}[!htbp]
\centering
\footnotesize
\caption{Comparison of mAP at different thresholds (0.5, 0.75, 0.9) for Single stage , Multi stage and transformer based detector with FVDNet with different backbone network on PASCAL VOC 2012 Dataset}
\begin{tabular}{lccc}
\hline
\textbf{Models/ mAP}                 & \multicolumn{1}{l}{\textbf{mAP@0.5}} & \multicolumn{1}{l}{\textbf{mAP@0.75}} & \multicolumn{1}{l}{\textbf{mAP@0.90}} \\ \hline
\multicolumn{4}{c}{\textbf{Multi Stage Detector}}                                                                                                           \\ \hline
\textbf{Fast R-CNN}                  & 0.66                                 & 0.53                                  & 0.26                                 \\ \hline
\textbf{Faster R-CNN v2}             & 0.76                                 & 0.61                                  & 0.32                                  \\ \hline
\multicolumn{4}{c}{\textbf{Single Stage Detector}}                                                                                                          \\ \hline
\textbf{Yolo v5}                     & 0.53                                    & 0.41                                     & 0.24                                    \\ \hline
\textbf{Yolo v6}                     & 0.68                                    & 0.44                                    & 0.25                                     \\ \hline
\textbf{Yolo v7}                     & 0.72                                    & 0.46                                     & 0.28                                     \\ \hline
\textbf{FVDNet}                & 0.67                                    & 0.41                                     & 0.23                                     \\ \hline
\textbf{FVDNet + R-152}   & 0.68                                    & 0.43                                     & 0.25                                     \\ \hline
\textbf{FVDNet + D-169} & 0.64                                    & 0.44                                     & 0.23                                    \\ \hline
\textbf{FVDNet + IN} & 0.62                                  & 0.45                                     & 0.24                                     \\ \hline
\multicolumn{4}{c}{\textbf{Transformer based Detector}}                                                                                                     \\ \hline
\textbf{Pix2Seq}                     & 0.65                                   & 0.45                                    & 0.24                                     \\ \hline
\end{tabular}
\label{tab:Tab5}
\end{table}

\subsection{Impact of changing the loss function}
In our experiments, we explored the use of Kullback-Leibler Divergence (KLD) as the loss function in combination with focal loss, replacing the previously used Jensen-Shannon Divergence (JSD). The results of FVDNet with different backbone architectures are presented in Table \ref{tab:Tab6}. Interestingly, the model achieved the highest mean Average Precision (mAP) across all thresholds when using the default backbone. However, when comparing these results with those obtained using JSD, we observed a degradation in precision.

\begin{table}[t]
\centering
\footnotesize
\caption{Results with KLD as the loss function for bounding box on FRUVEG67.}
\begin{tabular}{lccc}
\hline
\textbf{Models/ mAP}                 & \multicolumn{1}{l}{\textbf{mAP@0.5}} & \multicolumn{1}{l}{\textbf{mAP@0.75}} & \multicolumn{1}{l}{\textbf{mAP@0.90}} \\ \hline
\textbf{FVDNet}                & 0.74                                    & 0.62                                     & 0.33                                     \\ \hline
\textbf{FVDNet + R-152}   & 0.73                                    & 0.54                                    & 0.31                                     \\ \hline
\textbf{FVDNet + D-169} & 0.71                                    & 0.54                                     & 0.35                                     \\ \hline
\textbf{FVDNet + IN} & 0.71                                   & 0.51                                     & 0.22                                    \\ \hline
\textbf{FVDNet (JSD)}                & \textbf{0.78}                                    & \textbf{0.63}                                    & \textbf{0.37}                                     \\ \hline
\end{tabular}
\label{tab:Tab6}
\end{table}

In our experiments with the Faster R-CNN architecture, we also decided to investigate the impact of fixing the default variable grid size of the Region Proposal Network (RPN). We incorporated grid sizes of 8, 16, and 32 for generating anchors during the proposal stage. The results of these experiments are presented in Table \ref{tab:Tab7}, which demonstrates the performance on the FRUVEG67 dataset. We observed improvement of 1\% across different evaluation thresholds when using the fixed grid size approach. This suggests that for the FRUVEG67 dataset, smaller anchor boxes generated with the fixed grid size lead to more accurate detections. However, to gain a deeper understanding of the impact of the fixed grid size approach, we also evaluated its performance on the PASCAL VOC 2012 dataset, as shown in Table \ref{tab:Tab8}. Interestingly, on the PASCAL VOC 2012 dataset, we observed a decrease in accuracy compared to the variable grid size approach. These findings indicate that the impact of fixing the grid size of the RPN can vary significantly depending on the characteristics of the dataset. Therefore, the choice of grid size should be made with careful consideration of the dataset's object scales and other related factors.

\begin{table}[!htbp]
\centering
\caption{Impact of Fixed Grid Size of (32, 16 and 8) for Faster RCNN on FRUVEG67 with JSD as the bounding box loss.}
\begin{tabular}{lccc}
\hline
\textbf{Models/ mAP}     & \multicolumn{1}{l}{\textbf{mAP@0.5}} & \multicolumn{1}{l}{\textbf{mAP@0.75}} & \multicolumn{1}{l}{\textbf{mAP@0.90}} \\ \hline
\textbf{Faster R-CNN}    & 0.71                                 & 0.58                                  & 0.28                                  \\ \hline
\textbf{Faster R-CNN v2} & 0.77                                 & 0.62                                  & 0.32                                  \\ \hline
\end{tabular}
\label{tab:Tab7}
\end{table}

\begin{table}[!htbp]
\centering
\caption{Impact of Fixed Grid Size of (32, 16 and 8) for Faster RCNN on PASCAL VOC 2012 with JSD as the bounding box loss}
\begin{tabular}{lccc}
\hline
\textbf{Models/ mAP}     & \multicolumn{1}{l}{\textbf{mAP@0.5}} & \multicolumn{1}{l}{\textbf{mAP@0.75}} & \multicolumn{1}{l}{\textbf{mAP@0.90}} \\ \hline
\textbf{Faster R-CNN}    & 0.67                                 & 0.52                                  & 0.24                                  \\ \hline
\textbf{Faster R-CNN v2} & 0.71                                 & 0.54                                  & 0.29                                  \\ \hline
\end{tabular}
\label{tab:Tab8}
\end{table}

\section{Conclusion and Future Work}
\label{sec:conclu}
In conclusion, this research has made significant strides in the domain of fruit and vegetable detection in unconstrained environments. This research introduces the FRUVEG67 dataset, SSDA annotation, and the FVDNet model with JSD for improved fruit and vegetable detection in unconstrained environments. The potential applications of this research are noteworthy, particularly in the fields of electronics and agriculture. In electronics, the accurate detection and localization of fruits and vegetables can find application in automated sorting and packaging processes, contributing to the efficiency of food processing industries. Moreover, in agriculture, the developed methodologies can be employed for precision farming, aiding farmers in monitoring crop health, detecting diseases, and optimizing resource utilization. We also anticipate that the proposed FRUVEG67 dataset and the methodologies introduced herein will not only contribute significantly to the broader field of computer vision but also find practical applications in real-world scenarios, fostering advancements in electronics and agriculture.

 \bibliographystyle{elsarticle-num} 
 \bibliography{cas-refs}

\end{document}